\documentclass[conference]{IEEEtran}
\usepackage{subfigure}
\usepackage{graphicx}
\usepackage{multirow}

\ifCLASSINFOpdf
\else
\fi

\usepackage{amsmath}

\begin{document}
%
\title{Scene Text Eraser}


\author{\IEEEauthorblockN{Toshiki Nakamura\IEEEauthorrefmark{1},
Anna Zhu\IEEEauthorrefmark{2},
Keiji Yanai\IEEEauthorrefmark{3},and
Seiichi Uchida\IEEEauthorrefmark{1}}
\IEEEauthorblockA{\IEEEauthorrefmark{1}Human Interface Laboratory, Kyushu University, Fukuoka, Japan. Email: \{nakamura,uchida\}@human.ait.kyushu-u.ac.jp}
\IEEEauthorblockA{\IEEEauthorrefmark{2}School of Computer, Wuhan University of Technology, Wuhan, China. Email: annakkk@live.com}
\IEEEauthorblockA{\IEEEauthorrefmark{3}Department of Informatics, The University of Electro-Communications, Tokyo, Japan. Email: yanai@cs.uec.ac.jp}}

\maketitle

\begin{abstract}
The character information in natural scene images contains various personal information, such as telephone numbers, home addresses, etc. It is a high risk of leakage the information if they are published. In this paper, we proposed a scene text erasing method to properly hide the information via an inpainting convolutional neural network (CNN) model. The input is a scene text image, and the output is expected to be text erased image with all the character regions filled up the colors of the surrounding background pixels. This work is accomplished by a CNN model through convolution to deconvolution with interconnection process. The training samples and the corresponding inpainting images are considered as teaching signals for training. To evaluate the text erasing performance, the output images are detected by a novel scene text detection method. Subsequently, the same measurement on text detection is utilized for testing the images in benchmark dataset ICDAR2013. Compared with direct text detection way, the scene text erasing process demonstrates a drastically decrease on the precision, recall and f-score. That proves the effectiveness of proposed method for erasing the text in natural scene images.

\end{abstract}

%
\IEEEpeerreviewmaketitle

\section{Introduction}
\begin{figure}[b]
\centering
\subfigure[]{
\label{fig1_1}
\includegraphics[width=4.2cm,bb = 0 0 401 301]{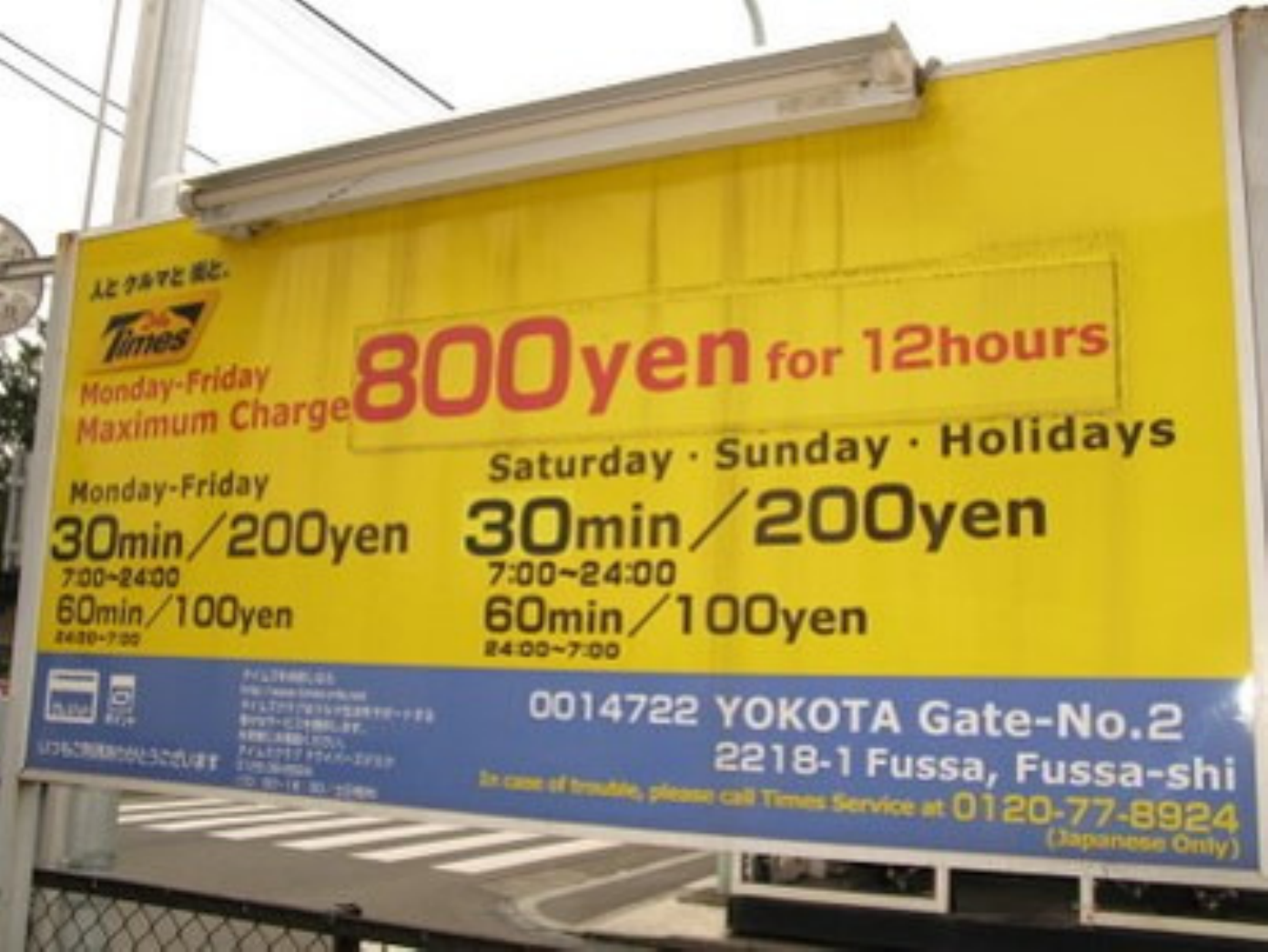}}
\subfigure[]{
\label{fig1_2}
\includegraphics[width=4.2cm,bb = 0 0 401 301]{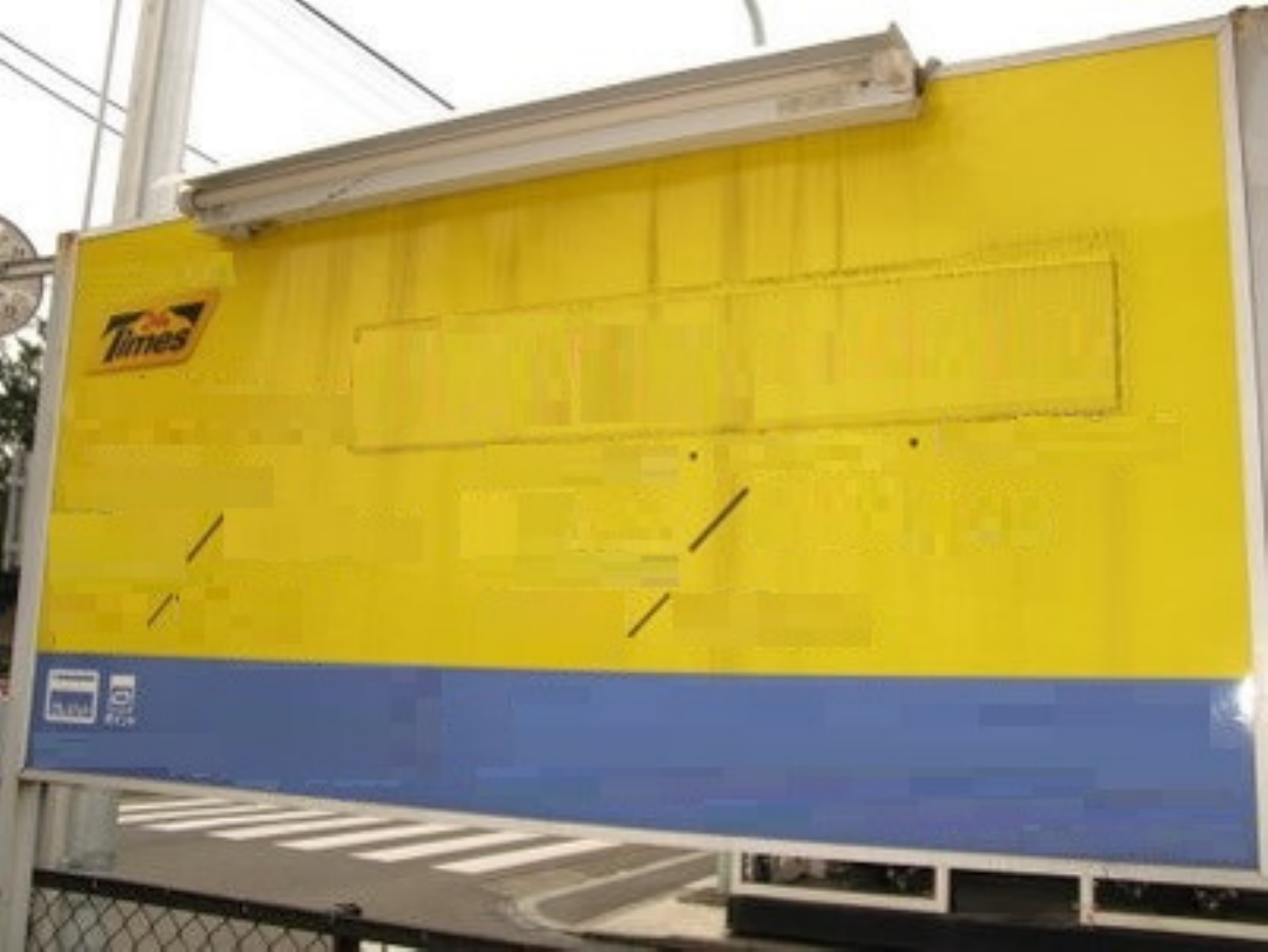}}
\caption{Hide text in scene images.}
\label{fig1}
\end{figure}

Nowadays, personal private information such as telephone numbers, ID number, home addresses, car numbers~\cite{inai2014selective}, etc. have become the special identity of person. Those important information may be incidentally captured, and appear in natural scene images. If published on the internet, it is a high risk to be collocated automatically by machines and criminals for illegal usage. To prevent the leakage of personal information, especially the text in scene images, information hidden technology is in great demand. Different from scene text detection~\cite{ye2015text}, the text hidden technics do not extract the whole text lines. That means perfect detection accuracy is not required. Only the characters or parts of them are removed and the process parts should not be distinct from the background. The example is shown in Fig.~\ref{fig1}.

The goal is to erase the text regions and make them hard to be detected. The simple image processing like blurring through Gaussian filter~\cite{carreiraperpinan2006fast} is only valid to text with specific shape and stroke. However, scene text has various appearance~\cite{jaderberg2016reading}, such as color, font, size, orientation, etc. Additionally, in the background, lots of clutters exist and effect the text and non-text judgement. Those challenges make the task difficult to solve. In this paper, we propose a novel method that erases the scene text via an inpainting deep neural network (DNN). 

The problem is converted as image transformation refereing to transforming images from a source image space to a target image space. In our case, it only needs input images and the output are text erased images with non-text regions remain original. The inpainting DNN is considered as the eraser. It composes of Convolutional neural networks (CNN) in front and deconvolutional neural networks (DeCNN)~\cite{noh2015learning} subsequently to recover the image resolution. The CNN is used to represent the feature of the image~\cite{yang2015multiple}. If only the features on the top are used for transformation, some details may be lost. To tackle this problem, interconnection between the deconvolutional layers and the convolutional layers which have the same size is built, and then the result is inputted to next deconvolutional layer. This model is trained in end-to-end fashion. We use inpainting~\cite{criminisi2004region} and dilation process to obtain the ground truth for training. 

A text detection method~\cite{liu2016ssd} then detects the text regions in the text erased images and the performance is evaluated in the same manner~\cite{karatzas2013icdar} including the precision, recall and f-score. Compared with the detection result on original ICDAR 2013 images, all the measurements decrease drastically on text erased images. That demonstrates effectiveness of our proposed method. 

The major contributions of our work are claimed as follows:
\begin{itemize}
\item We propose to use scene text eraser to hide text in the images without text detection. The concept is novel for preventing the leakage of private text-based information. And this scene text eraser can remove the text naturally and effectively.

\item The scene text eraser is implemented in image transformation way. The convolution-to-deconvolution structure adds the summation process among layers for better image quality.

\item The dealation and inpainting process are applied to label ground truth automatically and accurately.
\end{itemize}

The rest of the paper is structured as follows: A selection of related work is reviewed in Sect.~\ref{Sect2}. Sect.~\ref{Sect3} presents our proposed method in detail. In Sect.~\ref{Sect4}, we give the experimental results which include the details of databases and the experimental setup. Finally, Sect.~\ref{Sect5} gives a summarization and conclusion of this paper.

\section{Related work}~\label{Sect2}
Two strategies can be used for scene text erasing. One follws text detection pipelines~\cite{zhu2016scene,yin2014robust} that extract the text regions and then erase them by post-process. The other shares the idea of image transformation~\cite{oka1990method,johnson2016perceptual} that considers the output image as a different style in which the text are removed and the other parts keep original.

Generally, the text detection methods detect text through either connected component analysis (CCA)-based procedure or sliding window-based procedure. The CCA-based methods~\cite{huang2014robust,epshtein2010detecting} involves character candidates extraction, character/non-character classification, and text grouping. The sliding window-based methods~\cite{wang2012end,neumann2013scene} extract regional textual features, such as HoG, LBP~\cite{gan2011pedestrian}, CNN etc, from the regions which are scanned discretely from the image space by multi-scale and multi-ratio, and then scores the regions by inputting the features to a pertained text/non-text classification engine. Regions with high text scores are grounded to text regions eventually. Sometimes, image pre-processing or post-processing techniques are required and added in the two pipelines. For text erasing, further process is required, for instance, how to fill the text regions by background color.

In recent years, many classic problems can be framed as image transformation tasks~\cite{johnson2016perceptual}, where a system receives some input image and transforms it into an output image. Examples from image processing include denoising, super-resolution, and colorization, where the input is a degraded image (noisy, low-resolution, or grayscale) and the output is a high-quality color image. Examples from computer vision include semantic segmentation and depth estimation, where the input is a color image and the output image encodes semantic or geometric information about the scene. The related algorithms, either transfer the tone (color, contrast, saturation, etc.) of an image, preserving its patterns
and details, or distort the texture uniformly of an image to create ¡°style¡±. Scene text erasing can also be treated as a style transferring. Due to the richness of features that a deep CNN can possess, this task used to train a feedforward DNN in a supervised manner for transferring. Examples include the Ref~\cite{simo2016learning} that automatically converts complex rough sketches to line drawings, Ref~\cite{gatys2016image} converts the image style, Ref~\cite{iizuka2016let} performs color conversion on black and white images, etc. In this paper, we think out using image transform technology to hide the characters in the image by DNN with a special structure.

\section{Proposed method}~\label{Sect3}

\begin{figure}[t]
\centering
\includegraphics[width=8.8cm,bb = 0 0 1219 398]{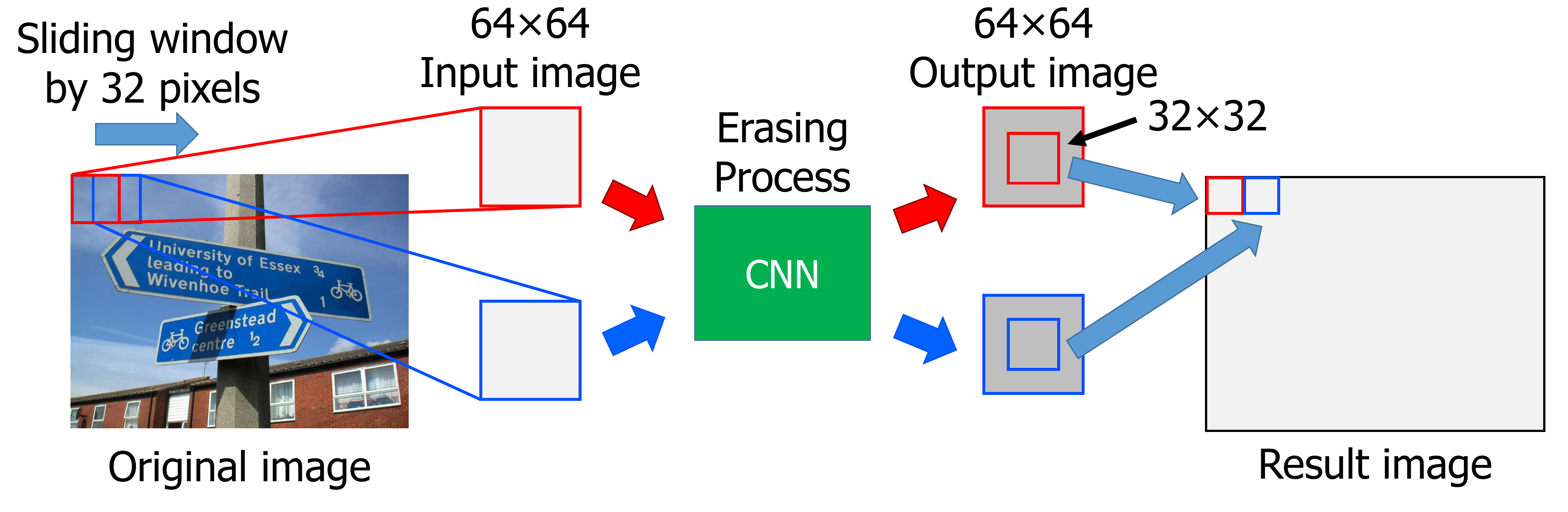}
\caption{The proposed method for scene text erasing.}
\label{fig2}
\end{figure}

The flowchart of the proposed method is shown in Fig.~\ref{fig2}. Since the purpose of text erasing is not the same as accurate text detection task, a single scale sliding window is applied to the original input images. The sliding stride is half of the window size. We cut the whole images into 64 $\times$ 64 patches and then input them into a pre-trained DNN. The size of each output result patches is also 64 $\times$ 64. To overcome the ambiguousness in the overlap regions, only the center part with 32 $\times$ 32 pixels of the output is considered valid and put back to original location. After this process, a single text hidden image is generated.

\subsection{The structure of the scene text eraser}

\begin{figure}[b]
\centering
\includegraphics[width=8.9cm,bb = 0 0 669 231]{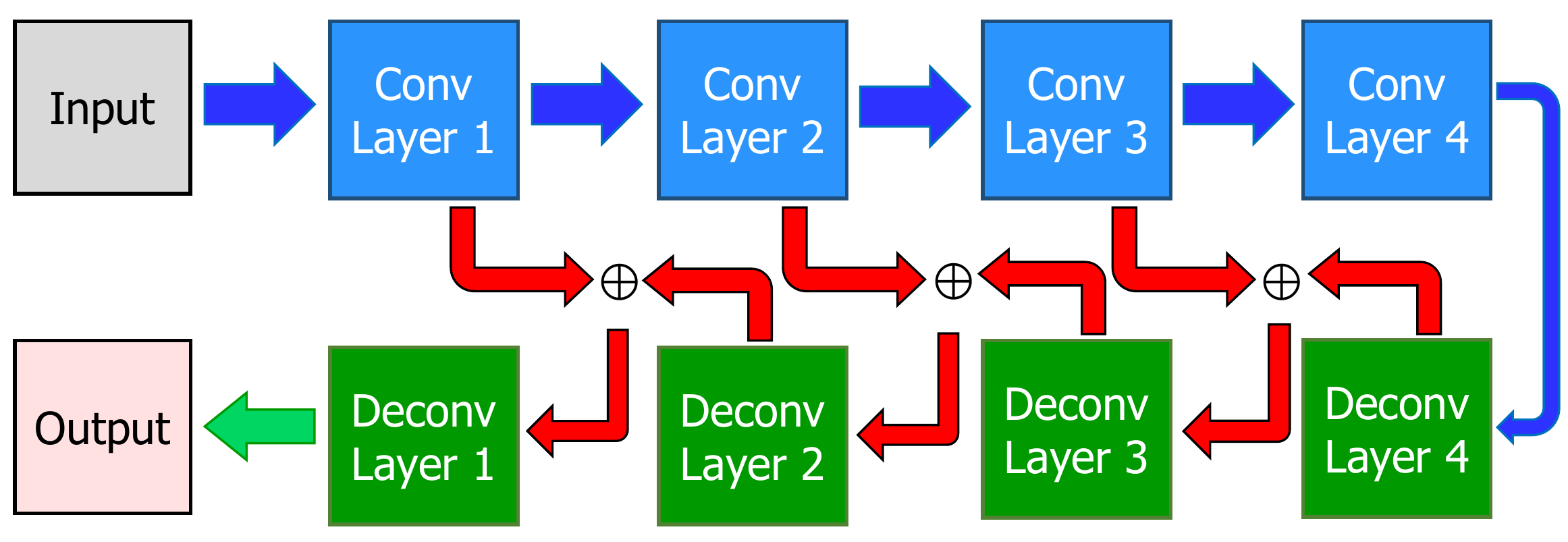}
\caption{The architecture of DNN in our proposed method.}
\label{fig3}
\end{figure}

A feedforward DNN composed with half convolution part and half deconvolution part is used as eraser in our approach. The architecture of the DNN is shown in Fig.~\ref{fig3}.

The convolution part contains four convolutional layers. The filter size of each convolutional layer is 4 $\times$ 4. The stride step and padding size is set to 2 and 1, respectively. Therefore, in each layer, the size of the feature maps reduces half comparing with the previous ones. The deconvolution part has the same structure but replaces the convolutional layers to deconvolution layer. The size of the filter, stride step and padding size is exactly the same as in convolution part. Thus, with the layer going deeper, the size of the feature maps is double increased. Due to the reduction of the image by convolution and the enlargement of the image by deconvolution, the output image has the same size as the original image.

However, if we only use a linear structure, in which the image size reduction or enlargement operations are performed isolated, lots of information on the original image may be lost. Because in the convolution part, only part of the information in the input image is stored, and the output image size is reduced. And in the deconvolution part, only the stored information is used to recover the image content. It results in information losing and low resolution of the output image.

To tackle this problem, we used skip connection technique~\cite{long2015fully} which is effective for restoring images with less deterioration. The skip connection sums the feature maps in different layer and inputs them to the next layer. Since the feature maps in convolution layers have more detailed information, such as the position information of objects, etc. By adding up the feautres of the previous layer for image recovering, it is possible to complement some image information that is lost by the reduction and enlargement procedure. And this process is expected to prevent the resolution being lowered. 

As shown in Fig.~\ref{fig3}, the skip connection is performed by adding a summation layer after each deconvolution layer. It is expressed in Eq.~\ref{eq1} by adding up the features $X_1$ in deconvolution layer and features $X_2$ in convolution layer element-wisely, and then inputing them to the next deconvolution layer. This summation layer requires the input from different layers have the same size. So the convolution to deconvolution structure is symmetry.

\begin{equation}
F(X_1,X_2) = \max(0,  X_1 + X_2).
\label{eq1}
\end{equation}

Rectified Linear Unit (ReLU)~\cite{nair2010rectified} is followed after each layer. Normalization is performed as well. Thus, the output result in each layer is rendered nonnegative. The lose function for back propagation uses mean square error~\cite{mao2016image} as expressed in Eq.~\ref{eq2}. $\emph{N}$ is the total training samples. $X_i$ represents the output through the skip connection DNN model and $Y_i$ is the text removed ground truth. We implement and train the network on Caffe. The stochastic gradient descent (SGD) with learning rate of $10^{-4}$ is used in training phase.

\begin{equation}
L(w) = \frac{1}{N}\sum_{i=1}^N\| F(X_i, w ) - Y_i\|^2
\label{eq2}
\end{equation}

\subsection{Training}

Since in our method, the patch images are input for DNN, we need to collect the training samples on patch level. The aim of the system is to hide the text information in natural scene images. So, for positive samples, the input are scene text images and the ground truth are the same images with text removed. For negative samples, the input and ground truth are the same background images. To automatically generate the training samples, the image inpainting process is performed. It is a technique to fill up defects in the images and make them inconspicuous. Specially, it is frequently used for restoring images when noises exist. For our case, the text is considered as defects and filled by the surrounding background color after inpainting.

\begin{figure}
\centering
\subfigure[]{
\label{fig5_1}
\includegraphics[width=4.2cm,bb = 0 0 247 187]{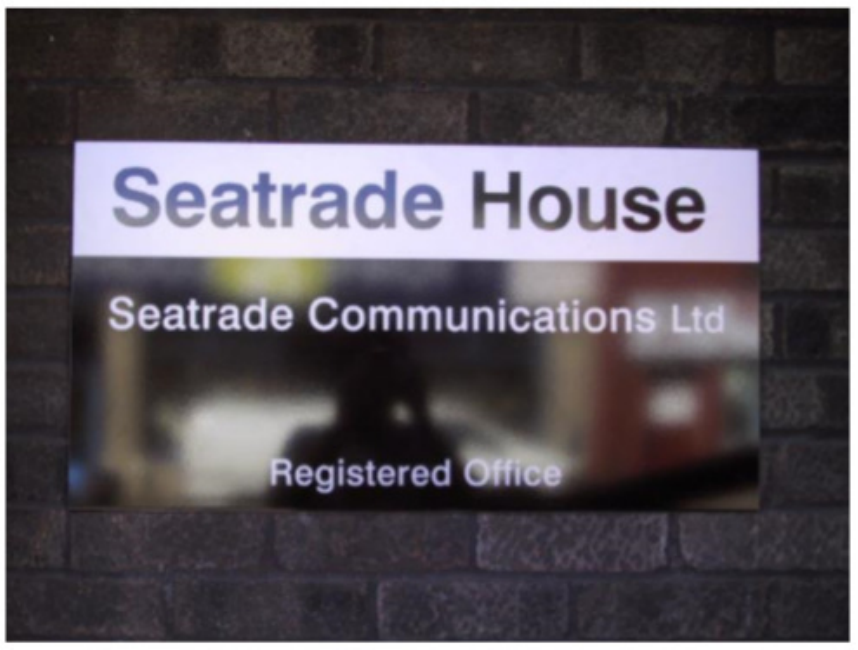}}
\subfigure[]{
\label{fig5_2}
\includegraphics[width=4.2cm,bb= 0 0 247 187]{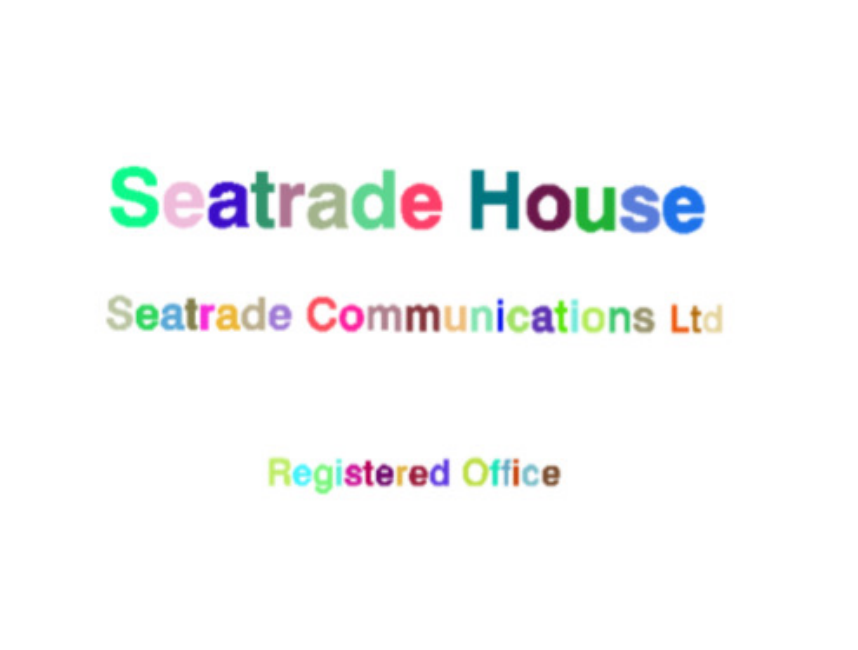}}
\subfigure[]{
\label{fig5_3}
\includegraphics[width=4.2cm,bb = 0 0 247 187]{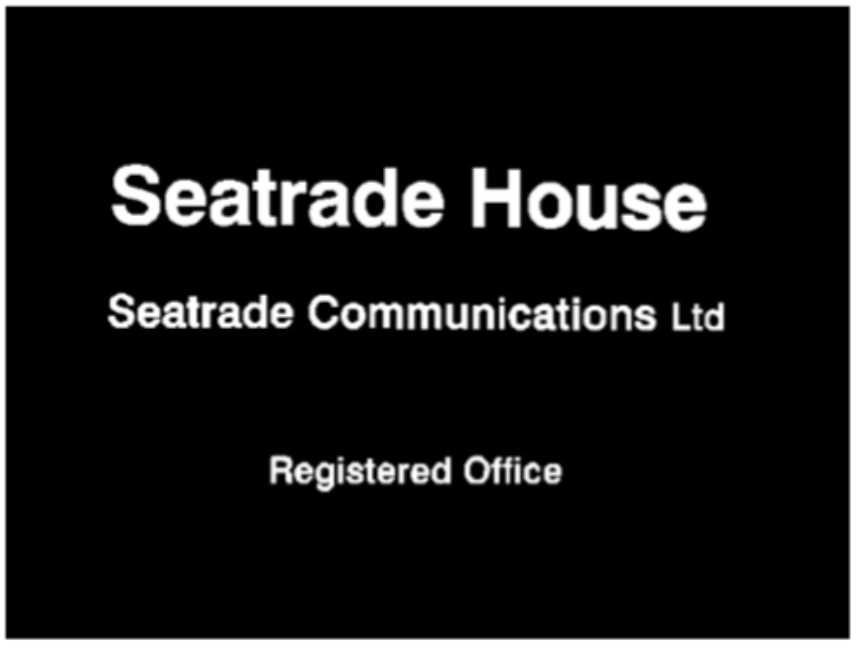}}
\subfigure[]{
\label{fig5_4}
\includegraphics[width=4.2cm,bb = 0 0 247 187]{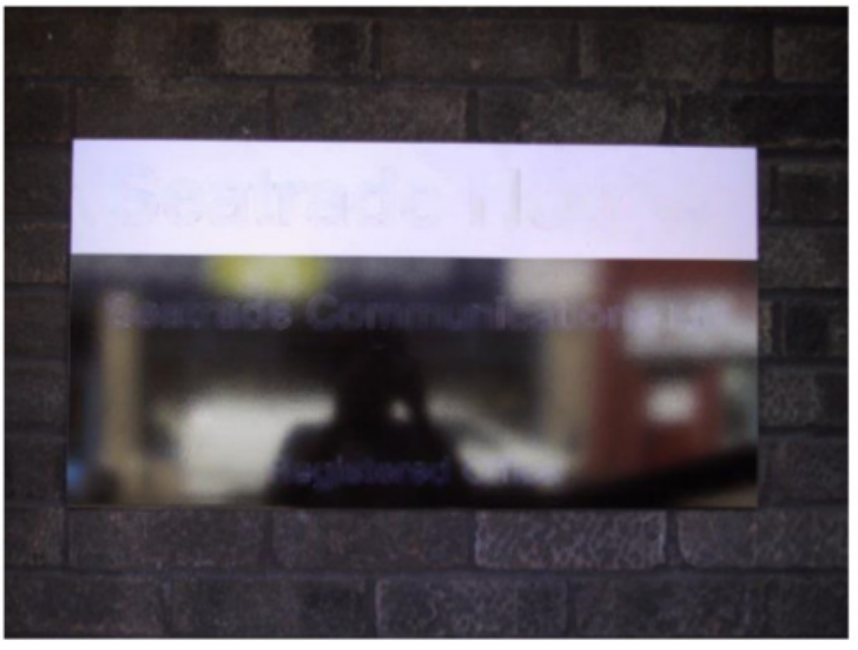}}
\subfigure[]{
\label{fig5_5}
\includegraphics[width=4.2cm,bb = 0 0 247 187]{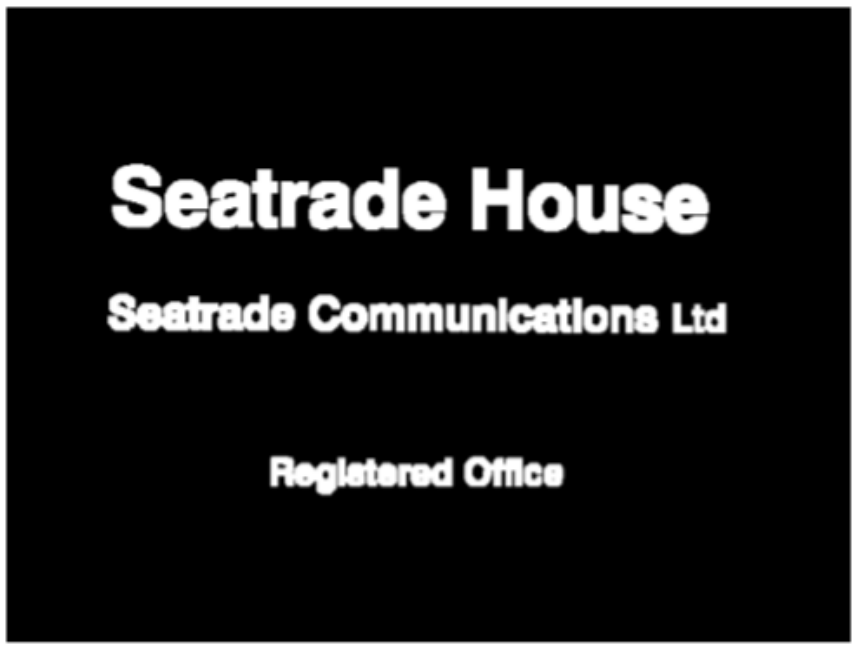}}
\subfigure[]{
\label{fig5_6}
\includegraphics[width=4.2cm,bb = 0 0 247 187]{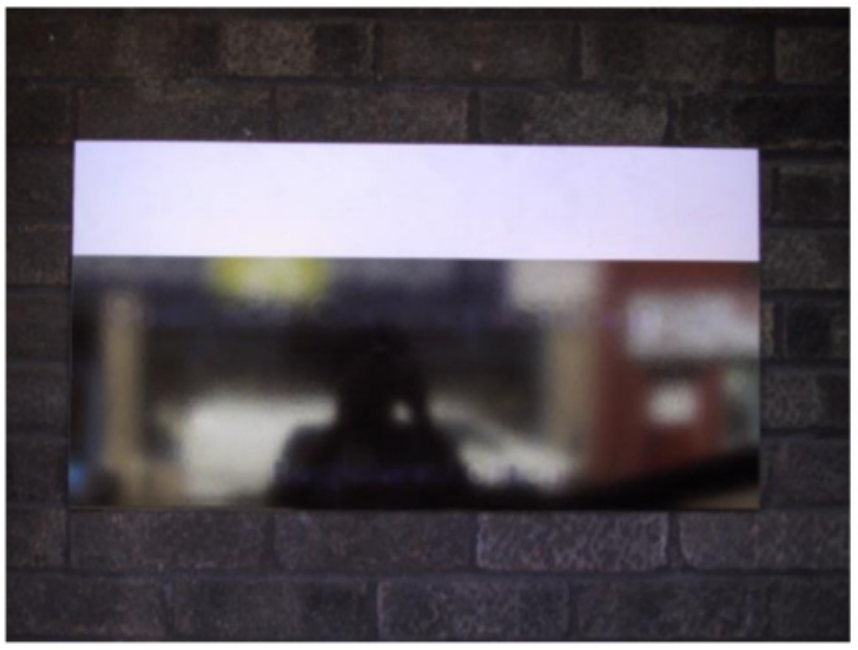}}
\subfigure[]{
\label{fig5_7}
\includegraphics[width=4.2cm,bb = 0 0 247 187]{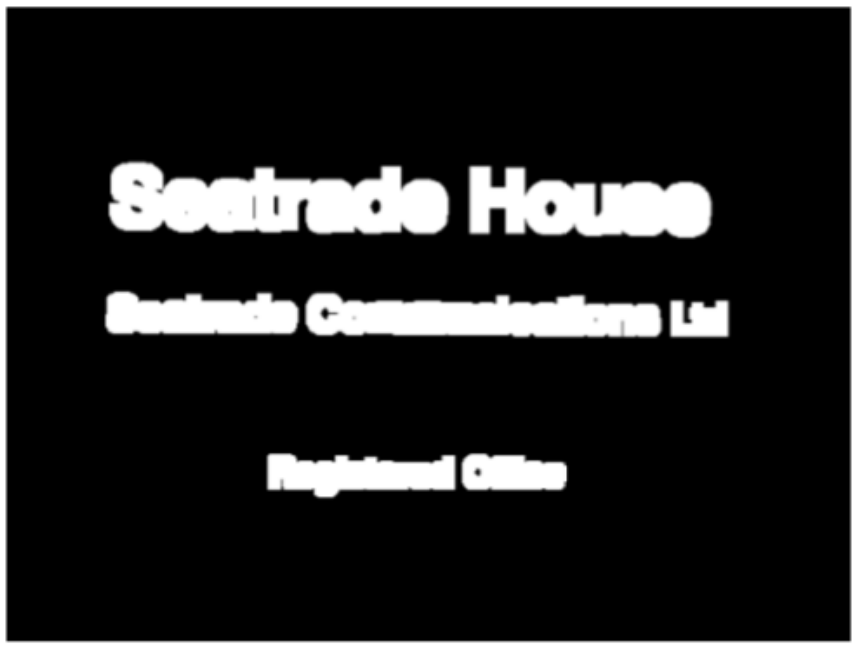}}
\subfigure[]{
\label{fig5_8}
\includegraphics[width=4.2cm,bb = 0 0 247 187]{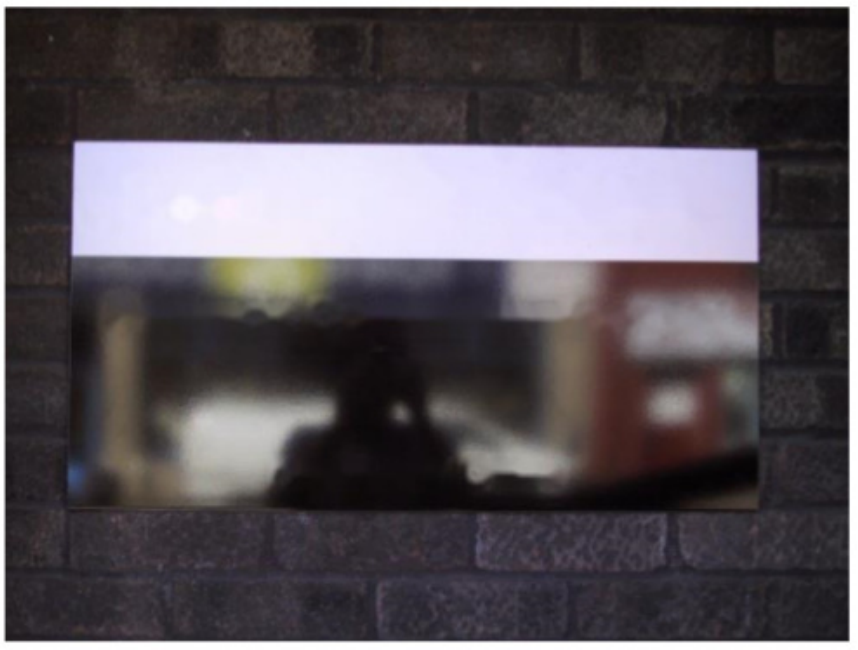}}
\caption{Ground truth generation. (a) Original image. (b) character level ground truth. (c) Binary character ground truth. (d) Inpainting result. (e) One time dilate result on binary character ground truth image. (f) Inpainting result based on binary image of (e). (g) Three times dilate result on binary character ground truth image. (h)Inpainting result based on binary image of (g).}
\label{fig5}
\end{figure}

Fig.~\ref{fig5} shows the details of the processing. Given the character ground truth in pixel level (Fig.~\ref{fig5_2}), inpainting process is applied on the original scene text image. The character ground truth is the basement as shown in Fig.~\ref{fig5_3}) and we can get the processing result in Fig.~\ref{fig5_4}. The pixels on character strokes are inpainted by the surrounding background color. To make the boundaries between character and background more inconspicuous in the image, additional dilation process is implemented on the basement images before image inpainting. Fig.~\ref{fig5_5} and ~\ref{fig5_7} is the dilate results by performing dilation once and three times, respectively. And Fig.~\ref{fig5_6} and ~\ref{fig5_8} is the final generation images by dilation and inpainting process sequentially.

\begin{figure}
\centering
\subfigure[]{
\label{fig4_1}
\includegraphics[width=8cm,bb = 0 0 792 292]{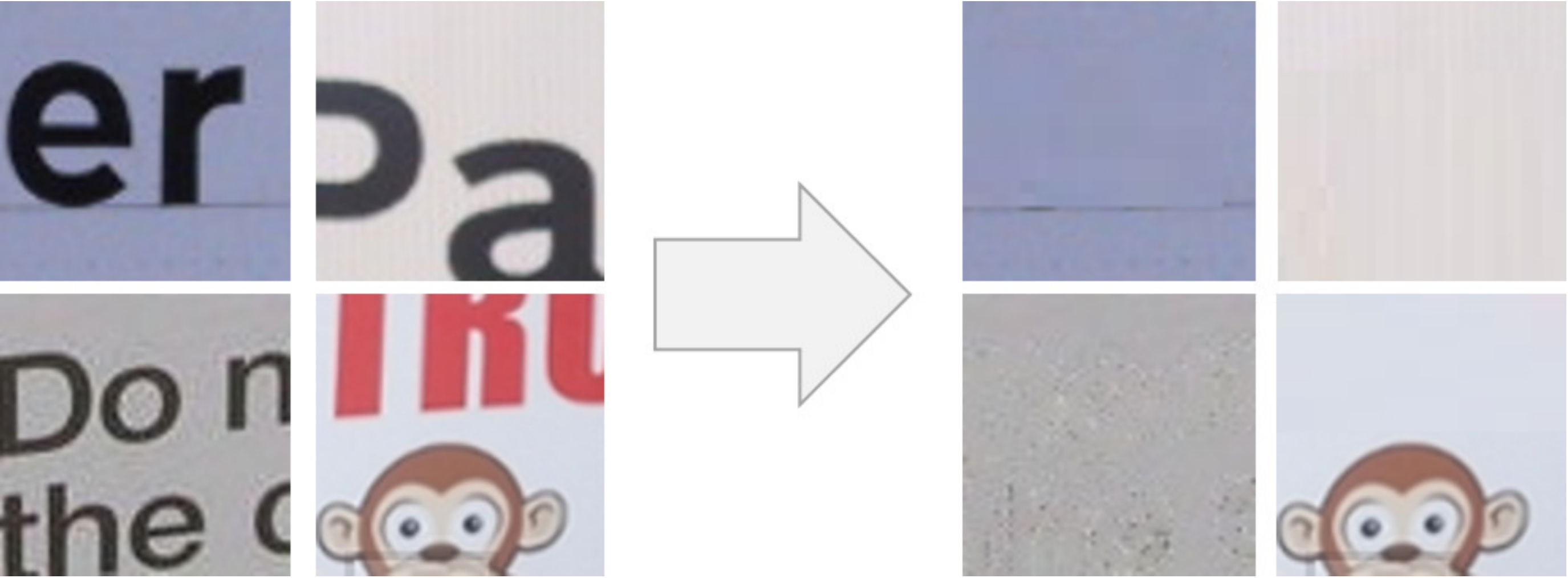}}
\subfigure[]{
\label{fig4_2}
\includegraphics[width=8cm,bb = 0 0 792 293]{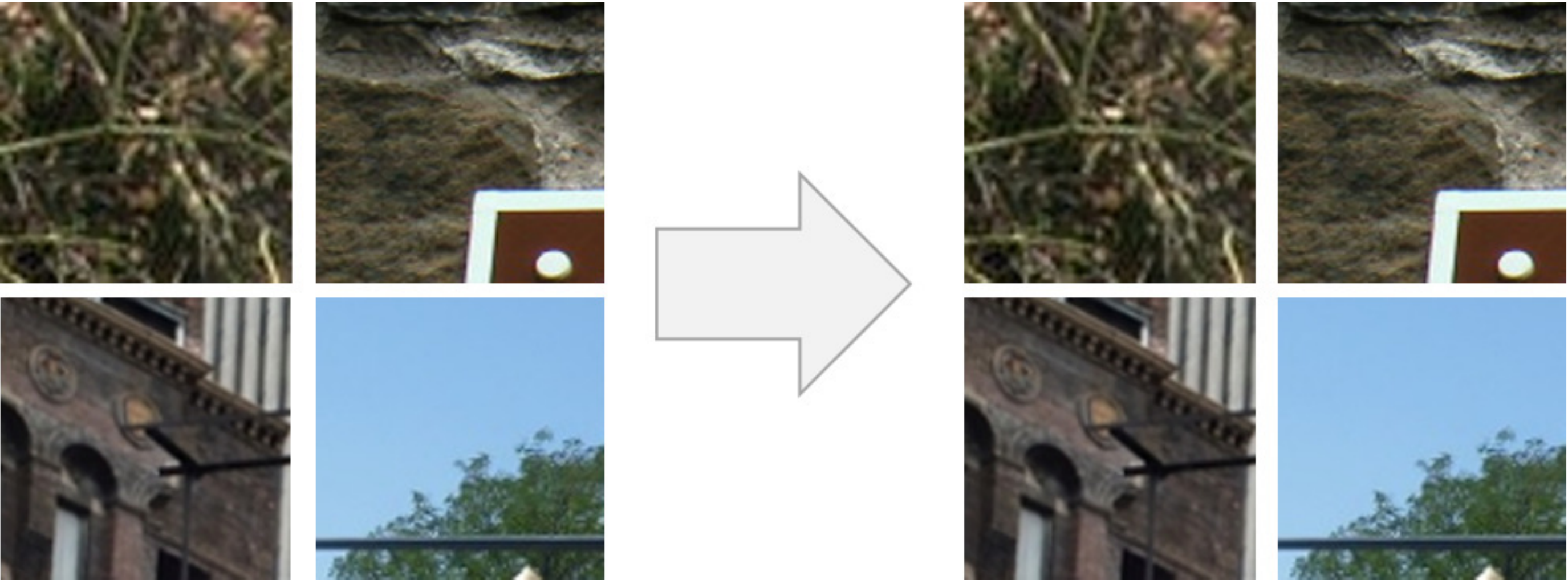}}
\caption{Examples of training samples for DNN learning.(a) Positive samples. (b) Negative samples.}
\label{fig4}
\end{figure}

To collect the patch level training samples, the sliding window with the size setting to 64 $\times$ 64 pixels is used. The batch formation is performed as well. The pair of input and output images are cropped from the same position in the original images and the corresponding inpainting images. The character ground truth is the guidance to classify the patches to positive or negative samples. In character ground truth images, if the corresponding cropped region contains any text, it is classified as text sample. Otherwise, that is background sample. Examples of the training data are shown in Fig.~\ref{fig4}. With this process, the training samples can be collected and classified automatically.

\section{Experimental results}~\label{Sect4}

\subsection{Dataset}
In the experiment, a Flickr image dataset which contains more than 3000 scene images and the benchmark dataset ICDAR 2013~\cite{karatzas2013icdar} which contains 229 images used for training. Most of the images in this dataset have signboards and billboards with text attached on. The font, color and position of characters and the background is various which is benefit for training the model. To evaluate the performance, the dataset ICDAR 2013 that is different from images used in training is tested. 

\subsection{Qualitative Evaluation}
\begin{figure}
\centering
\subfigure[]{
\label{fig6_1}
\includegraphics[width=8.9cm,height=10.5cm,bb= 0 0 368 394]{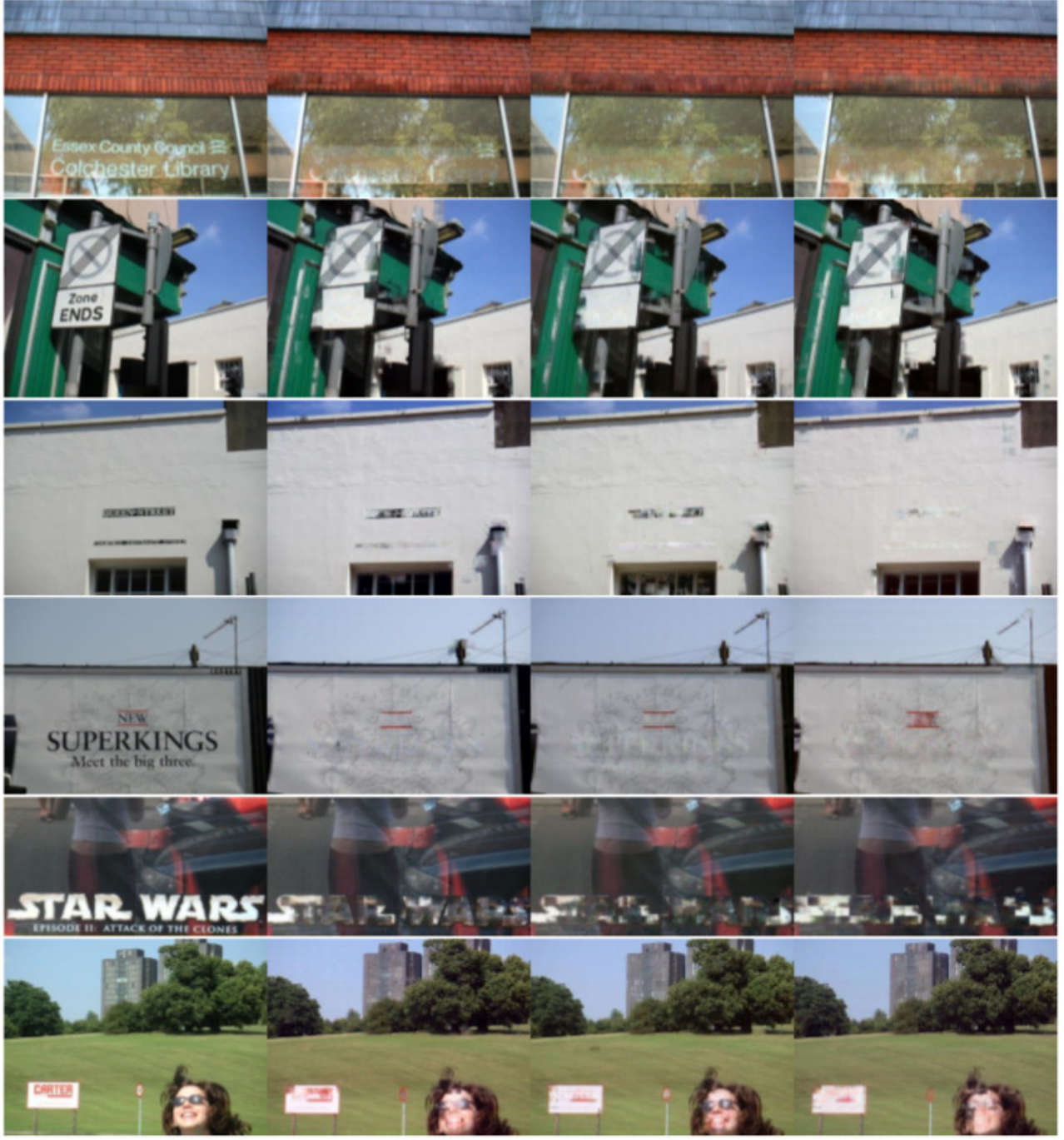}}
\subfigure[]{
\label{fig6_2}
\includegraphics[width=8.9cm,bb = 0 0 367 416]{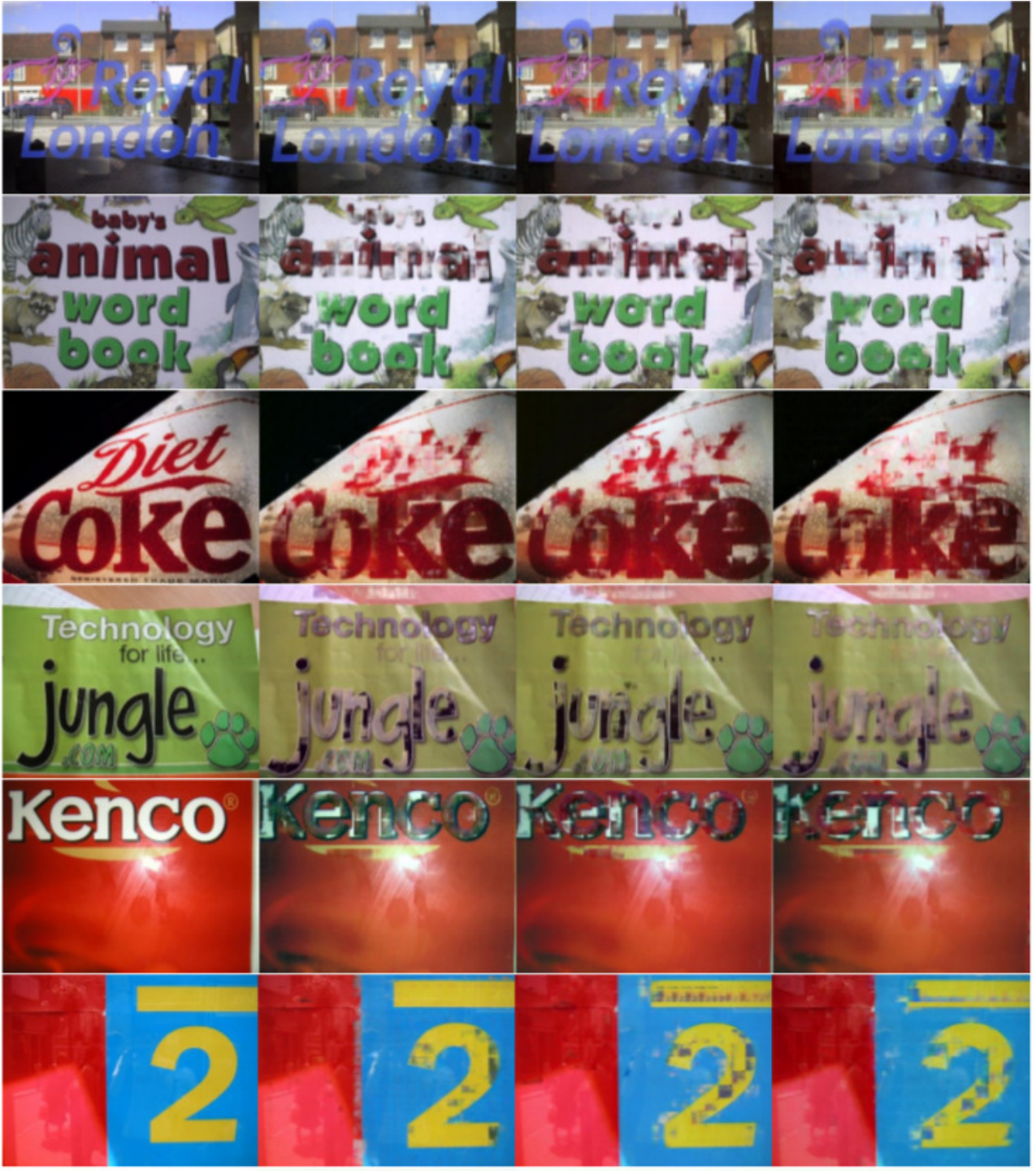}} 
\caption{Examples of text erased images. Images in the columns from left to right correspond to the original images, text erased images by training with only inpainting ground truth, text erased images by training with one time dilatation and inpainting ground truth, text erased images by training with three times dilatation and inpainting ground truth.}
\label{fig6}
\end{figure}

Fig.~\ref{fig6} shows some text erased image by employing our proposed method. In Fig.~\ref{fig6_1}, text can be successfully erased, even they are in complicated background, such as the the glass, the trees, etc. However, our proposed model fails for some cased as shown in Fig.~\ref{fig6_2}. Our work only uses one scale sliding window to get the subregion. The captured parts in the character whose size is much larger than the window size might be considered as background. So the output of the DNN has no changes in that subregion. This results in the bad erasing performance on images with large size characters. Comparing results from differently trained DNN,  the one that is trained with three times dilatation and inpainting ground truth gets the best performance. As shown in Fig.~\ref{fig6}, text in the images of the last column are mostly erased and can not be distinguished by human. Since the dilate operation turns more pixels on the character boundary to be considered as part of the character, the text erasing result looks smoothing and natural.

\subsection{Quantitative Evaluation}
To evaluate the scene text erasing performance, a modified text detection method~\cite{liu2016ssd} is used to detect the text in images after erasing process. It is an object proposal based deep neural network that predicts discrete regions with different aspect ratios and scales from multiple feature maps. To make it adapted for text detection, we select six aspect ratios: 0.7, 1, 2, 3, 5, 7 for designing the default boxes. The scales on the prediction layers range from 0.06 to 0.85. In total, 38124 regions estimated. Most of them are non-text regions. Only the detections with text probability higher than 0.7 are remained as text. We test this text detection method on scene text erased images in ICDAR 2013 and compare the results with original scene text images. They are named by the generation ways as below:

\begin{itemize}
\item Original images dataset: focused scene text images in ICDAR 2013.

\item Erased0 images dataset: Scene text erased images of ICDAR 2013 by network trained with inpainting ground truth.

\item Erased1 images dataset: Scene text erased images of ICDAR 2013 by network trained with one time dilatation and inpainting ground truth.

\item Erased3 images dataset: Scene text erased images of ICDAR 2013 by network trained with three times dilatation and inpainting ground truth.

\end{itemize}

\begin{figure}[t]
\centering
\includegraphics[width=8.9cm,bb = 0 0 568 1109]{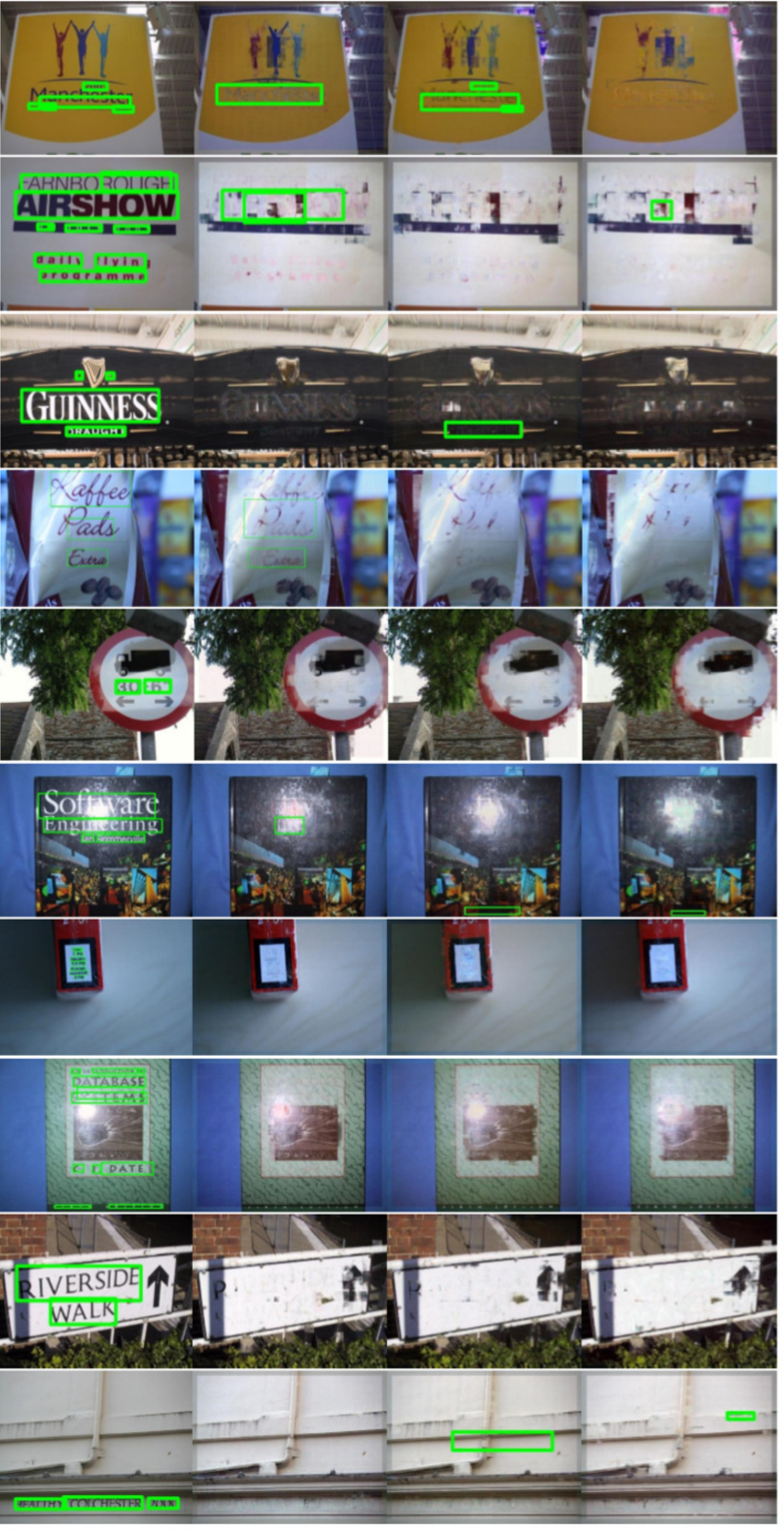}
\caption{Text detection performance on original images and text erased images. The detect results in the columns from left to right correspond to Original images dataset, Erased0 images dataset, Erased1 images dataset and Erased3 images dataset.}
\label{fig7}
\end{figure}

We follow the text detection performance measurement by compute the precision, recall and f-score under two protocols, the DetEval~\cite{wolf2014evaluation} and the ICDAR 2013 evaluation~\cite{karatzas2013icdar}. Precision represents the proportion of detected text regions to all detected regions. Recall is the proportion of detected text regions to ground truth text regions. f-score is a trade-off between precision and recall rate by computing their harmonic mean. Table~\ref{table1} demonstrates the results. After text erasing, the recall of text detection decreases more than 70\%. That demonstrates less text regions are detected. The precision decreases about 30\% representing that the non-text regions' proportion becomes higher in all the detected regions. Compared with the text detection results on original images, the three text erased image datasets have worse performance. The overall measurement f-score drops drastically after text erasing in the images. Inversely, it proves the effectiveness of the proposed method. As explained above, by adding the dilate operation for training samples, the text can be erased more smoothly and naturally. Without the shape boundaries between background and text regions, the erased text regions are much difficult to be detected. Examples of text detection results are displayed in Fig.~\ref{fig7}. The proposed text eraser can distinguish the text regions and non-text regions well. From the results, we can see that most text regions go through exserting process and are hidden afterwards. 

In this work, we only used single scale sliding window-based method to perform text erasing in images. It has some weakness for erasing large size text. In our future work, a real end-to-end system will be employed. The input is a complete scene text image, and the output is the text erased image. For training, the full images and the corresponding inpaining images will be the training samples instead of using cropped image patches. Additionally, we will propose new evaluation method to measure the character erased performance but not only by text detection evaluation.

\begin{table*}
\renewcommand{\arraystretch}{1.3}
\caption{The text detection performance on four datasets.}
\label{table1}
\centering
\begin{tabular}{|c||c|c|c|c|c|c|}
\hline
\multirow{2}{*}{Image dataset}& \multicolumn{3}{|c|}{ICDAR Eval} & \multicolumn{3}{|c|}{DetEval}\\

\cline{2-7} & Recall&  Precision & f-score & Recall&  Precision & f-score \\
\hline
Original images & 82.56\% & 83.70\% & 83.13\% & 81.90\%& 87.15\% & 84.45\%\\

Erased0 images & 21.74\%& 69.31\% & 33.09\%  & 22.25\%& 70.17\% & 33.78\%\\

Erased1 images & 13.88\%& 59.20\% & 22.49\% & 14.45\%& 60.48\% & 23.32\%\\

Erased3 images & 8.35\% & 54.07\% & 14.46\% & 8.89\%& 54.53\% & 15.30\%\\
\hline
\end{tabular}
\end{table*}

\section{Conclusion}~\label{Sect5}
To protect privacy of the text based information in natural scene images, we proposed a novel scene text eraser. It used the image transform method which transferred the scene text images to text erased images via an inpainting deep neural network.  This network process the image patches, which are cropped by sliding window, from convolution to deconvolution. To improve the resolution of output images and conserve more information of the non-text part in the original images, we used skip connection to sum the feature maps in both deconvolutional layers and specified convolutional layers. For model training, the dilate and inpainting technologies are applied subsequently to generate the training samples. A text detection method evaluated the text erasing performance on ICDAR 2013 dataset. The precision, recall and f-score dropped drastically after erasing the text in images. It proved the effectiveness of this text eraser. In our future work, we will develop this model in end-to-end fashion and think out new evaluation method to better measure the performance of scene text eraser. 

\section{Acknowledgments}
The pictures of left bottom of Fig.5(a) and left top of Fig.5(b) are taken from Flickr under the copyright license.
The authors would like to thank the contributors of those pictures.
Left bottom of Fig.5(a) and Left top of Fig.5(b) : alykat



 
%

\end{document}